\documentclass[final]{IEEEtran}

\usepackage{amsmath,amsfonts}
\usepackage{algorithmic}
\usepackage{algorithm}
\usepackage{array}
\usepackage[caption=false,font=normalsize,labelfont=sf,textfont=sf]{subfig}
\usepackage{textcomp}
\usepackage{stfloats}
\usepackage{url}
\usepackage{verbatim}
\usepackage{cite}
\usepackage{multirow}
\usepackage{graphicx}
\usepackage{float}
\usepackage[table]{xcolor}
\usepackage{hyperref}
\graphicspath{{Figures/}}
\DeclareGraphicsExtensions{.pdf,.png,.jpg, .svg}

\usepackage{balance}

\begin{document}

\title{Unsupervised Domain Adaptation for Semantic Segmentation using One-shot Image-to-Image Translation via Latent Representation Mixing}

\author{Sarmad F. Ismael, Koray Kayabol, and Erchan Aptoula

\thanks{Sarmad F. Ismael and Koray Kayabol are with the Electronics Engineering Department, Gebze Technical University, 41400 Kocaeli, Türkiye (e-mail:sismeal@gtu.edu.tr, sarmad.fakhrulddin88@gmail.com;koray.kayabol@gtu.edu.tr).} \thanks{Erchan Aptoula is with the Computer Science \& Engineering Department, Sabanci University, 34956 Istanbul, Türkiye (e-mail: eaptoula@sabanciuniv.edu).}}

\markboth{}
{Shell \MakeLowercase{\textit{et al.}}: A Sample Article Using IEEEtran.cls for IEEE Journals}
\maketitle

\begin{abstract}
Domain adaptation is one of the prominent strategies for handling both domain shift, that is widely encountered in large-scale land use/land cover map calculation, and the scarcity of pixel-level ground truth that is crucial for supervised semantic segmentation. Studies focusing on adversarial domain adaptation via re-styling source domain samples, commonly through generative adversarial networks, have reported varying levels of success, yet they suffer from semantic inconsistencies, visual corruptions, and often require a large number of target domain samples. In this letter, we propose a new unsupervised domain adaptation method for the semantic segmentation of very high resolution images, that i) leads to semantically consistent and noise-free images, ii) operates with a single target domain sample (i.e.~one-shot) and iii) at a fraction of the number of parameters required from state-of-the-art methods. More specifically an image-to-image translation paradigm is proposed, based on an encoder-decoder principle where latent content representations are mixed across domains, and a perceptual network module and loss function is further introduced to enforce semantic consistency. Cross-city comparative experiments have shown that the proposed method outperforms state-of-the-art domain adaptation methods. Our source code will be available at \url{https://github.com/Sarmadfismael/LRM_I2I}.
\end{abstract}

\begin{IEEEkeywords}
Unsupervised domain adaptation, semantic segmentation, latent representation mixing, one-shot learning, image translation. 
\end{IEEEkeywords}

\section{Introduction}
\IEEEPARstart{S}{upervised} semantic segmentation networks have been extensively explored for land use/land cover map production purposes from remote sensing images and great advances have been reported in this regard \cite{Peng22}. However, their large-scale application is hindered by the distribution discrepancy or shift, between source and target domains, usually stemming from geographical, temporal, and/or sensor based acquisition differences, and almost always leads to significant performance degradation \cite{Zhang22}.

Certainly, one could always retrain a model using target domain samples. But unfortunately, semantic segmentation networks require pixel-level labels that are notoriously time-consuming and expensive to obtain \cite{Zheng22}. Unsupervised domain adaptation on the other hand, can address this issue by transferring the knowledge learned from labelled source domain samples to the target domain through the exploitation of unlabelled target domain samples \cite{Tuia16}.

\begin{figure}[t]
  \begin{center}
  \includegraphics[width=0.35\textwidth]{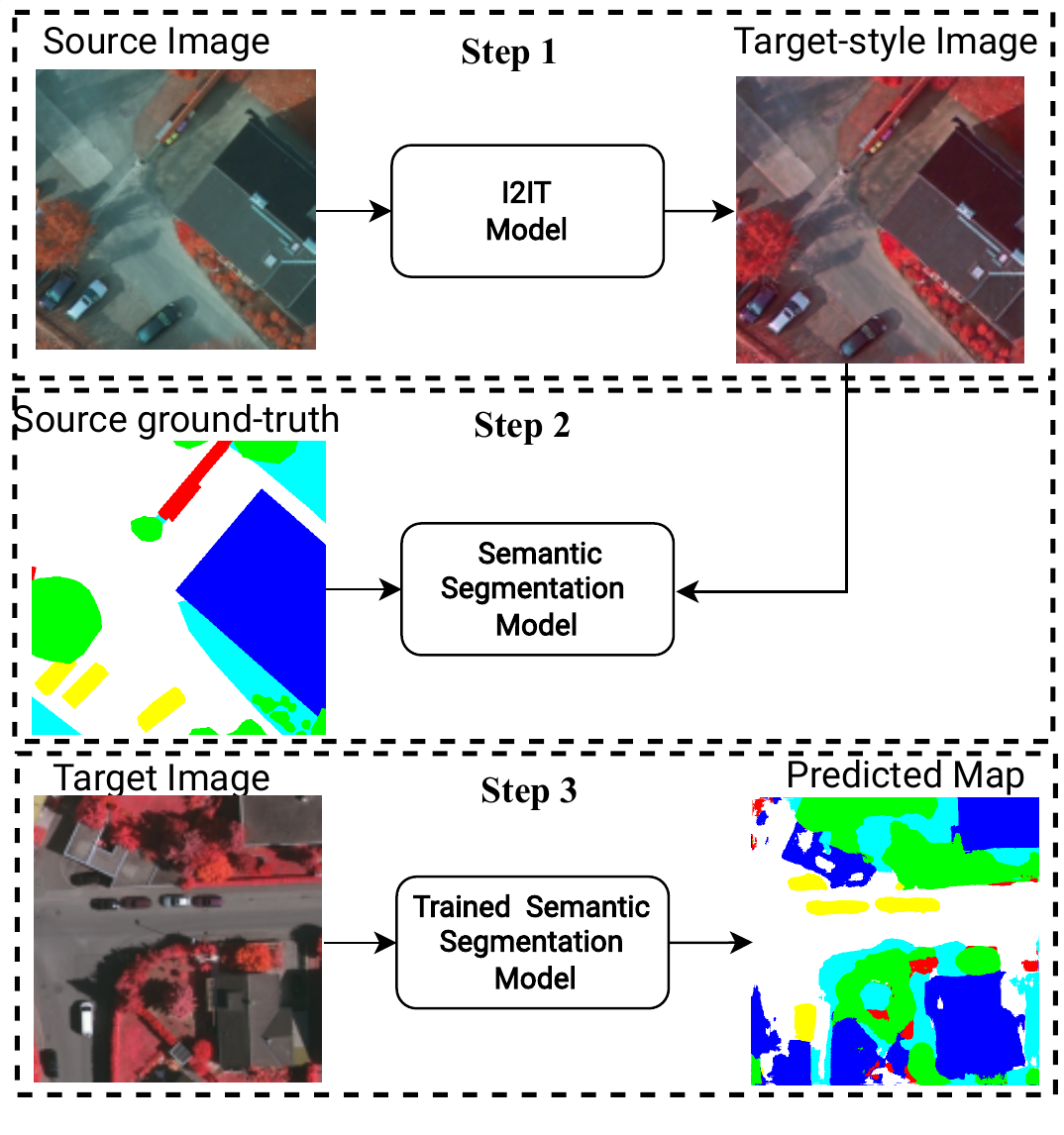}
  \end{center}
  \caption{Outline of the overall unsupervised I2I translation based method.}
  \label{fig1}
\end{figure}

To this end, a large variety of deep domain adaptation approaches have been explored, which can be classified into three categories: namely \textit{discrepancy based}, \textit{reconstruction based} and \textit{adversarial} \cite{Csurka2017}. Discrepancy based methods aim to directly minimize the feature distribution difference between source and target domains using various first order (e.g.~maximum mean discrepancy) \cite{Liu20} and second order metrics (e.g.~correlation alignment) \cite{Wang21}. Next, in reconstruction based approaches, the target domain reconstruction acts as an auxiliary task to promote domain invariance through a shared latent representation across domains \cite{Miao21}. And last, adversarial approaches either encourage domain confusion via an adversarial goal against a discriminator \cite{Liu22}, or employ generative adversarial networks (GANs) to generate source-domain images in the style of the target domain (Image to Image Translation - I2IT) in order to promote domain invariant features \cite{Mateo21}. 

Due to its often superior performance, the generative adversarial research direction has been explored intensively; e.g.~CycleGAN 
\cite{Benjdira19} has been a groundbreaking contribution leveraging both the GAN concept and the cycle-consistency loss function for I2IT. DualGAN \cite{Li21} constitutes another step in this direction, where two parallel GANs are exploited in order to generate new target-style images. ColorMapGAN \cite{Tasar20}, on the other hand, has relied on the concept of color and spectrum translation in order to restyle the source domain samples.

\begin{figure*}[t]
\centering
\includegraphics[width=\linewidth ]{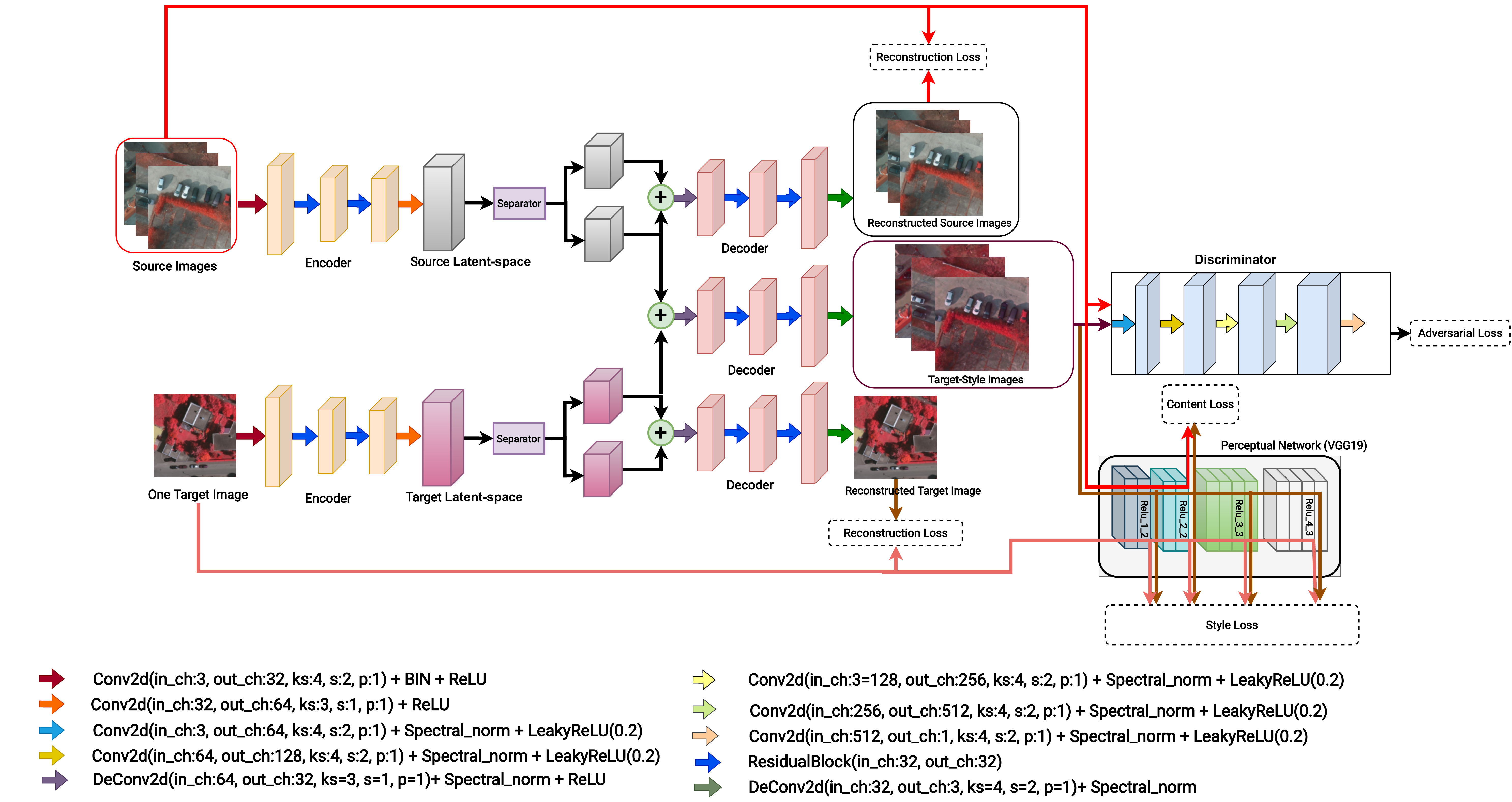}
\caption{Overall block diagrams of the proposed I2IT.}
\captionsetup{justification=centering}
\label{fig2}
\end{figure*}

Although the aforementioned approaches have led to significant advances, they suffer from several drawbacks; for instance, semantic inconsistencies are known to appear when the domain shift is relatively large \cite{Benjdira19}, while synthesized images can often be corrupted \cite{Li21}. Moreover, GANs are infamously known for their high training sample demand \cite{Zhu17}.

To address these shortcomings, a new I2IT network model is proposed in this letter. Contrary to the state-of-the-art, 
\begin{enumerate}
\item it avoids the complexity of GANs, and is based on an encoder-decoder principle relying on latent representation separation and mixing across domains and is equipped with a perceptual network module and loss in order to promote semantic and visual consistency,
\item it is able to operate at a competitive level even with \textit{a single target domain sample}, thus placing it in the efficient one-shot learning category, reducing training computational cost
\item it offers this performance while requiring only a fraction of the network parameters of state-of-the-art methods.
\end{enumerate}
The proposed approach has been validated through the ISPRS Potsdam and Vaihingen very high spatial resolution datasets, where it has outperformed several state-of-the-art methods.

\section{PROPOSED METHOD}
The unsupervised domain adaptation context is used when one has access to source and target domain data, in addition to only source domain labels, and has to learn a model that predicts the labels of the target domain data. The main idea (Fig.~\ref{fig1}) consists of synthesizing target-style source images that can then be used to train a semantic segmentation network via the source domain's labels. The method's novelty lies in the design of the I2IT module (Fig.~\ref{fig2}), relying on cross-domain content mixture through an encoder-decoder structure, in addition to the use of a perceptual network module resulting in content and style losses.

The proposed I2IT module consists of four types of components: encoders, separators/combiners, decoders, and two loss modules: discriminator and perceptual networks. 
The encoder and decoder components are employed during both the training and testing phases to discover latent representations and then synthesize the target-styled images, whereas the discriminator and perceptual loss networks are only employed in the training phase in order to optimize the objective function. 

\subsection{Latent-space Separation}
 A formal definition of content and style remains elusive, for this purpose, we adopt the method of \cite{Gatys15,Johnson16}, and assume that the content vector representing the semantic layout (i.e.~scene structure) can be extracted using the higher layer of an encoder network. To preserve semantic consistency while minimizing noise, the proposed I2IT method separates the last layer feature maps constructed by the encoder, which represents the content of the source ($I_s$) and target images ($I_t$) into two halves, and then synthesizes a new target-style image ($I_{s\to t}$) by combining the first half of the source domain sample with the second half of the target domain sample. 

As shown in Fig.~\ref{fig2} the two encoders ($E$) map the source and target images to two latent-spaces ($\mathcal{L}_s$ and $\mathcal{L}_t$). By means of the separator component ($Sep$), the latent feature is separated into two halves: first half and second half:
\begin{equation} \label{eq1}
\mathcal{L}_{s} = E\left (I_{s} \right ), \; \left [\mathcal{L}_{s1},\:\mathcal{L}_{s2}  \right ] = Sep(\mathcal{L}_{s}).
\end{equation}
\begin{equation} \label{eq2}
\mathcal{L}_{t} = E\left (I_{t} \right ),\; \left [\mathcal{L}_{t1},\:\mathcal{L}_{t2}  \right ] = Sep(\mathcal{L}_{t}).
\end{equation}
The decoders ($D$) on the other hand reconstruct three images: the reconstructed source image ($I_{s}^{'}$) obtained by combining the source first half and source second half; the target-styled source image $I_{s\to t}$, representing the desired new synthesized image, by combining the source first half and target second half; and finally, the reconstructed target image ($I_{t}^{'}$) by combining the target first half and target second half. These can be denoted as:
\begin{equation}\label{eq3}
I_{s}^{'} = D\left (\mathcal{L}_{s1},\:\mathcal{L}_{s2}  \right ).
\end{equation}
\begin{equation}\label{eq4}
I_{s\rightarrow t} = D\left (\mathcal{L}_{s1},\:\mathcal{L}_{c2}  \right ).
\end{equation}
\begin{equation}\label{eq5}
I_{t}^{'} = D\left (\mathcal{L}_{c1},\:\mathcal{L}_{c2}  \right ).
\end{equation}
Furthermore, the discriminator ($Dis$) has been introduced to distinguish between the $I_s$ and the $I_{s\to t}$. Moreover, inspired by \cite{Johnson16}, the perceptual network ($P$) represented by the VGG19 model pre-trained on the ImageNet dataset has been used in order to evaluate and thus improve the preservation of the content and style of the $I_{s\to t}$. 

\subsection{Loss Functions}
In order to construct the objective functions of the proposed I2IT networks, four loss functions have been used (i.e.~the dotted boxes in Fig.~\ref{fig2}): reconstruction loss ($L_{Rec}$), adversarial loss ($L_{Adv}$), and perceptual loss ($L_C$ and $L_S$). 

$L_{Rec}$ ensures that the encoder and decoder are a pair of opposite mappings to one another by using the $L_1$ distance between the original and the reconstructed images:
\begin{equation}\label{eq6}
L_{Rec}(E,D) = \left \| I_{s} - I_{s}^{'} \right \|_{1} +  \left \| I_{t} - I_{t}^{'} \right \|_{1}.
\end{equation}

The purpose of $L_{Adv}$ is to enforce the indistinguishability of the synthesized target-style image from the source domain samples by means of the discriminator networks. A modified version of the standard GAN form called LSGAN \cite{ref17} has been employed to this end:
\begin{equation}
\begin{split}
L_{Adv}(Dis) = \mathbb{E}_{I_{s}\sim^{P_{data}(I_{s})}} \left [ \left ( Dis(I_{s}) - 1 \right )^{2} \right ] + \\
\mathbb{E}_{I_{s\rightarrow t}\sim^{P_{I_{s\rightarrow t}}(I_{s\rightarrow t})}}\left [ \left ( Dis(I_{s\rightarrow t}) \right )^{2} \right ].
\end{split}
\end{equation}\label{eq7}
\begin{equation}
L_{Adv}(E,D) =\mathbb{E}_{I_{s\rightarrow t}\sim^{P_{I_{s\rightarrow t}}(I_{s\rightarrow t})}}\left [ \left ( Dis(I_{s\rightarrow t}) - 1 \right )^{2} \right ].
\end{equation}
The perceptual loss function includes content loss $L_C$ and style loss $L_S$, the use of $L_C$ was inspired from \cite{Gatys15} and its purpose is content preservation across source and target-style samples. It is calculated by taking the squared L2 norm between the feature produced by the perceptual network $P$ at a given layer $i$, which is represented by the output of ($ReLU2\_2$) layer:
\begin{equation}
L_{C}(E,D) = \sum_{i}\left \| P_{i}(I_{s}) - P_{i}(I_{s\rightarrow t}) \right \|_{2}^{2}.    
\end{equation}
$L_S$ on the other hand aims to minimize the style loss between the target and newly synthesized target-style image. It is dependent on the squared Frobenius norm between the Gram matrices of the features produced by the perceptual network $P$ at layer $j$, which correspond to the outputs of ($ReLU1\_2$, $ReLU2\_2$, $ReLU3\_3$, $ReLU4\_3$) layers:
\begin{equation}
 L_{S}(E,D) = \sum_{j}\left \| G_{j}(I_{t}) - G_{j}(I_{s\rightarrow t}) \right \|_{F}^{2}.
\end{equation}
Let the shape of the feature representation map of an image ($I$) at the output of layer $j$ of network $P$ be $Ch \times H \times W$; where $Ch$ is the number of channels, $H$ its height and $W$ its width. In which case, the Gram matrix $G_{j}$ is the matrix with element at $ch, ch'$: 
\begin{equation}
G_{j}^{ch,ch^{'}}(I)= \frac{1}{Ch_{j}H_{j}W_{j}} \sum_{h=1}^{H_j}\sum_{w=1}^{W_j} P_{j}^{h,w,ch}(I) P_{j}^{h,w,ch^{'}}(I). 
\end{equation}

Consequently the overall objective function becomes: 
\begin{equation}
    L_{Total}(E,D) = \lambda _{1} L_{Rec}\;+\;  \lambda _{2} L_{Adv}\;+\;\lambda _{3} L_{c}\;+\;\lambda _{4} L_{s}\:
\end{equation}
where $\lambda_1$, $\lambda_2$, $\lambda_3$, $\lambda_4$ are regularization parameters. 

\section{EXPERIMENTS}
\subsection{Dataset}
In order to test the proposed unsupervised domain adaptation method, the ISPRS 2D semantic segmentation benchmark dataset \cite{ref18} has been employed. It contains very high resolution aerial images of two distinct geographic locations in Germany; namely Potsdam, consisting of 38 samples with a resolutions of 5 cm/px and a size of $6000 \times 6000$ px, and Vaihingen, consisting of 33 samples with a resolution of 9 cm/px and a size of approximately $2000 \times 2000$ px. All images have been acquired with 3 channels (as near infrared, red, green - IR-R-G) and its ground truth is fully available at pixel-level, with six classes: namely clutter/background (BA), building (BU), low vegetation (LV), tree (TR), car (CA), and impervious surfaces (IS). All images were cropped into $512 \times 512$ px patches, leading to 4598 and 1798 samples from Potsdam and Vaihingen respectively. They have been split into training/validation/testing randomly with a 70/15/15 ratio. 

\begin{table*}[ht]
\caption{Unsupervised domain adaptation results across Potsdam and Vaihingen in both directions.}
\begin{center}

\resizebox{18cm}{!}
	{\scriptsize
	\begin{tabular}{|c|c|c|c|c|c|c||c|c|c|c|c|}
	\hline
	\multicolumn{12}{|c|}{\textbf{Potsdam} $\rightarrow$ \textbf{Vaihingen (P2V)}}\\
	\hline
	\hline
	\multirow{2}{*}{Methods} & \multicolumn{6}{c}{mIoU} & \multicolumn{3}{||c|}{Average}&\multirow{2}{*}{Sec/iteration} &\multirow{2}{*}{\# of parameters}  \\    
	\cline{2-10}
	& BA & BU & LV & TR & CA & IS & mIoU & F1 & OPA & &  \\
	\hline
	Lower baseline & 0.75 & 50.71 & 9.17 & 26.86 & 1.97 & 10.52	 &  16.66  &  25.14  & 33.38 & - & - \\
   \hline
  CycleGAN \cite{Zhu17} & \textbf{12.57} &  36.91 & 26.08 &  42.78 & 18.35 & 41.76 & 29.74 & 44.58 & 53.33 & 0.531 & 11.378M \\ 
  \hline
  Benjdira et al.~\cite{Benjdira19}& 8.88 & 49.89 & \textbf{27.98} & 35.01 & 22.85 & \textbf{44.15} & 31.46 & 46.15 & 56.39 & 0.531 & 11.378M  \\ 
  \hline
  DualGAN \cite{Li21} & 1.91 &  35.51 & 6.73 &	43.15 &	 7.06 &	6.5 & 16.81 & 25.74 & 39.1 & 0.207 & 41.828M\\
  \hline
  ColorMapGAN \cite{Tasar20} & 10.12 &	 49.33 & 25.37 & 36.6 & 2.41 &	24.03 & 24.64 & 36.99 & 51.24 & \textbf{0.054} & 100.663M \\ 
    \hline
  Proposed &  6.88  & \textbf{53} & 16.48 & \textbf{58.9} & \textbf{23.72} & 43.13 & \textbf{33.69}  &  \textbf{47.13} & \textbf{60.09} & 0.16 & \textbf{0.115M} \\ 
  \hline
   Upper baseline & 26.18 & 88.76 & 65.22 & 74.51 & 61.98 & 79.66 & 66.05 & 77.52 & 86.85  & - & - \\ 
    \hline
     \hline
   \multicolumn{12}{|c|}{\textbf{Vaihingen} $\rightarrow$ \textbf{Potsdam (V2P)}}\\
	\hline
	\hline
	Lower baseline & 5.49 & 39.93 & 46.99 & 6.39 & 19.9 & 30.65 & 24.89 & 37.26 & 51.19  & - & - \\
    \hline
      CycleGAN \cite{Zhu17} & 2.95 &	\textbf{45.35} &	 33.83 & 31.81 & 0.01 & \textbf{47.92} & 26.98 & 38.63 & \textbf{56}& 0.531 & 11.378M \\ 
    \hline
     Benjdira et al.~\cite{Benjdira19}& \textbf{5.06} & 42.57 & 28.57 & \textbf{33.19}  & 16.76  & 40.15  & 27.72  & 41.6  & 52.27& 0.531 & 11.378M \\ 
    \hline
     DualGAN \cite{Li21} & 2.27 & 	34.15 &	 44.35 & 21.48 & 28.95 & 37.58 & 28.13 & 41.95 & 51.54& 0.207 & 41.828M \\
     \hline
    ColorMapGAN \cite{Tasar20} & 2.42 & 17.39 & 35.46 &	1.22 & 0.44 & 37.97 & 15.82 & 24.17 & 43.74 & \textbf{0.054} & 100.663M\\ 
    \hline
    Proposed & 2.79 &  41.51 &  \textbf{42.55} & 29.28 &  \textbf{39.27 } & 42.91 &  \textbf{33.05}	&  \textbf{47.46} & 55.74 & 0.16 & \textbf{0.115M} \\
    \hline
    Upper baseline & 45.78 & 86.21	& 67.67	 & 66.17 & 74.7	& 77.75	& 69.71 & 81.46 & 84.41  & - & - \\ 
    \hline

    \end{tabular}}
 \end{center}
\label{results}
\end{table*}

\begin{figure*}[h]
\begin{center}
\includegraphics[width=0.9\textwidth ,height=5cm]{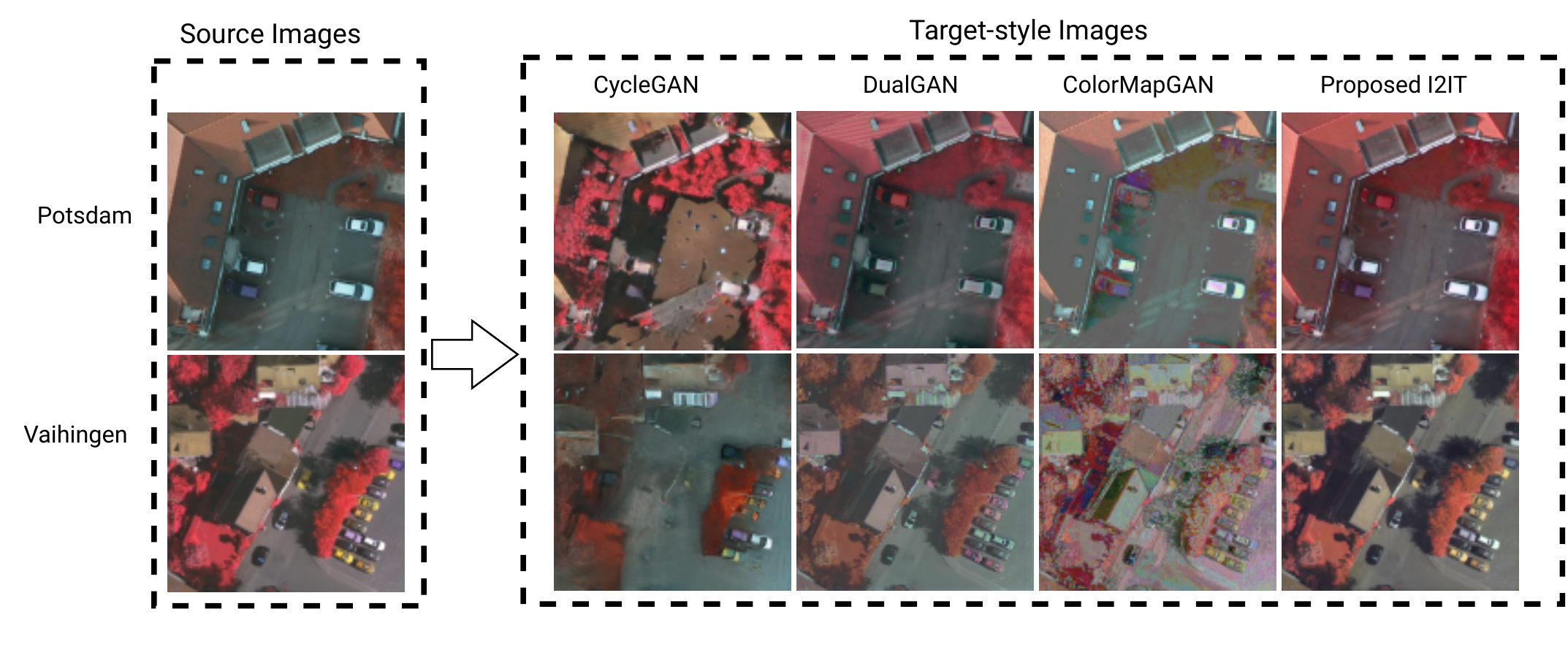}
\end{center}
\caption{Target-style images output of source images (Potsdam, Vaihingen) dataset using CycleGAN \cite{Zhu17}, DualGAN \cite{Li21}, ColorMapGAN \cite{Tasar20}, and the proposed I2IT approach.}
\label{fig3}
\end{figure*}
\subsection{Setup}
The details of the I2IT network are provided in Fig.~\ref{fig2}. The batch instance normalization (BIN) has been used to accelerate the training steps. Spectral normalization \cite{ref20} has been applied in the decoder and discriminator to make training more stable. The $\mathrm{\lambda}_1$, $\mathrm{\lambda}_2$, $\mathrm{\lambda}_3$, and $\mathrm{\lambda}_4$ in the total objective function were set to 30, 1000, 1, and 5, respectively. As to the semantic segmentation model, the basic U-Net model \cite{ref1} has been used with ResNet101 as the encoder backbone with weight initialization based on ImageNet.

All components of the proposed method are implemented using the Pytorch framework on a NVIDIA RTX 3060 GPU with 12GB of memory. The optimization method is Adam, with an initial learning rate of 0.001, weight decay of 0.0005, and beta values of (0.5, 0.999). The models have been trained for 100 epochs with a batch size of 8 using an early stopping criterion based on the validation loss. Performance has been measured using the mean intersection over union (mIoU), F1-score, and overall pixel accuracy (OPA).

Two cross-domain adaptation scenarios have been explored: Potsdam to Vaihingen (P2V) and vice versa (V2P). The proposed approach has been compared against four counterparts: CycleGAN \cite{Zhu17}, Benjdira et al.~\cite{Benjdira19}, DualGAN \cite{Li21}, and ColorMapGAN \cite{Tasar20}. To train the I2IT model, we employ only a single target sample with the proposed approach, assumed to represent all classes. In order to mitigate the effect of the chosen sample bias, the experiments have been repeated with 10 different samples, and the mean performances have been reported. The other methods employ all the available target domain samples.

\subsection{Results}
Fig.~\ref{fig3} shows the visual results of target-style synthesis. CycleGAN performed well transferring the style of source images to the target image, especially for the vegetation class. However, it removed some small objects such as cars and replaced them with other large objects such as buildings and the background. The transferred images generated by the DualGAN appear to be more realistic. Yet, the target-style resolution is reduced  (i.e.~the output is blurred). The ColorMapGAN on the other hand produces noisy output with all classes of the target-style image as it transforms each image channel separately. The proposed method generates semantically consistent target-style images that appear to be noise-free. 
\begin{figure*}[h]
\centering
\includegraphics[width=0.9\linewidth]{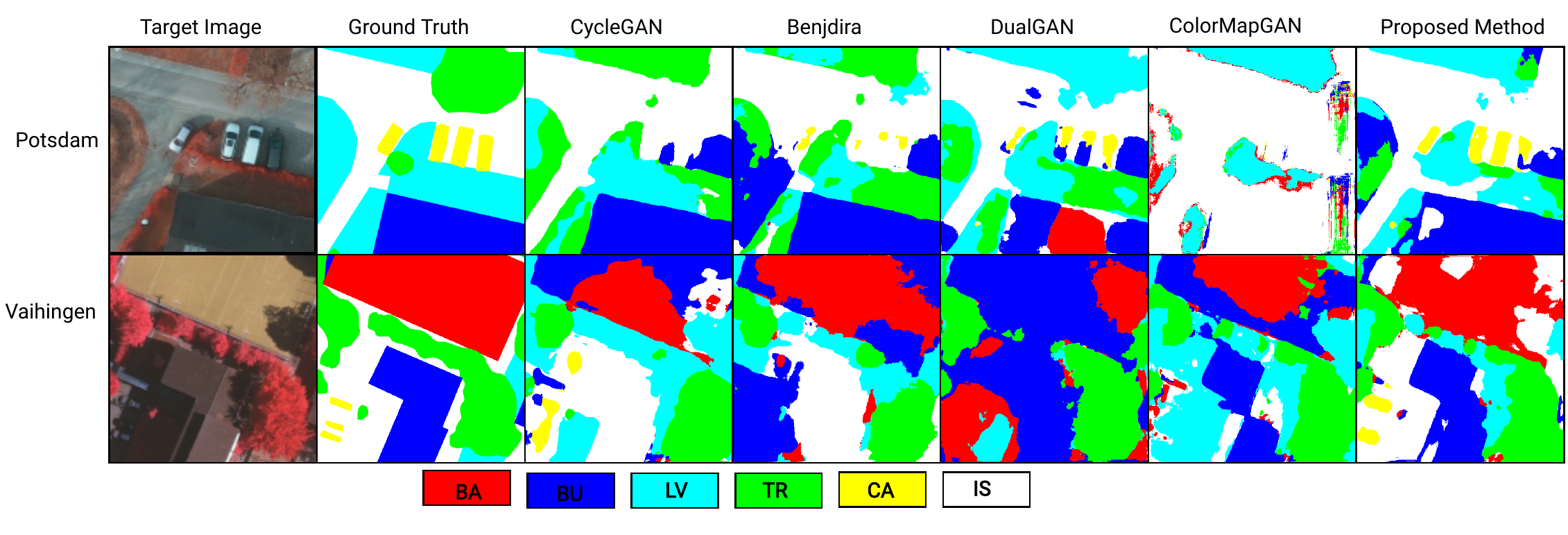}
\caption{The prediction maps of the target image for both cities (Potsdam, Vaihingen) obtained using CycleGAN \cite{Zhu17}, Benjdira et al.~\cite{Benjdira19}, DualGAN \cite{Li21}, ColorMapGAN \cite{Tasar20}, and the proposed approach.}
\captionsetup{justification=centering}
\label{fig4}
\end{figure*}

Table \ref{results} provides the quantitative adaptation results. The lower baseline (i.e.~no adaptation) represents the lower performance. The upper baseline is computed by training and testing the semantic segmentation model on the target domain. In the V2P scenario, the various tested domain adaptation methods, have all enhanced segmentation performance w.r.t.~the lower baseline except for ColorMapGAN. The proposed method in particular has outperformed in average across classes, all other methods w.r.t~all three performance metrics and both scenarios. However this superiority is not consistent across all classes. CycleGAN for instance achieved a higher performance than the proposed method in the background class during the P2V case and the building and impervious surfaces classes during the V2P case. The modification of the CycleGAN method proposed in \cite{Benjdira19} has also scored high values in the low vegetation class during the P2V case and the background and trees classes in the V2P case. This could be attributed to the (visually successful) color-wise appearance change of these classes post-translation. Moreover, the last two columns provide a measure of computational cost as seconds per iteration and the total number of parameters for training the proposed I2IT model. It is shown that the proposed method takes a shorter time than CycleGAN \cite{Zhu17} and DualGAN \cite{Li21} but is more complex than ColorMapGAN \cite{Tasar20}. Furthermore, the number of parameters is significantly lower w.r.t.~all counterparts.

Fig.~\ref{fig4} additionally shows the predictions maps obtained across the tested methods, where one can observe the relatively higher accuracy of the region borders computed by the proposed approach.

\section{CONCLUSION}
In this letter a new I2IT method is proposed to be used in the context of unsupervised domain adaptation for the semantic segmentation of very high resolutions remote sensing images. The presented method is based on the combination of a latent-space separation with a perceptual loss in order to overcome the semantic inconsistency issue widely encountered in the state-of-the-art. Furthermore, it operates with a single target domain image for the sake of efficiency, while the experiments have shown consistently higher performance w.r.t.~its counterparts. Future work will focus on learnable separation, cross-sensor and temporal adaptation.

\bibliographystyle{IEEEtran}
\bibliography{bibliography.bib}

\begin{thebibliography}{10}
\providecommand{\url}[1]{#1}
\csname url@samestyle\endcsname
\providecommand{\newblock}{\relax}
\providecommand{\bibinfo}[2]{#2}
\providecommand{\BIBentrySTDinterwordspacing}{\spaceskip=0pt\relax}
\providecommand{\BIBentryALTinterwordstretchfactor}{4}
\providecommand{\BIBentryALTinterwordspacing}{\spaceskip=\fontdimen2\font plus
\BIBentryALTinterwordstretchfactor\fontdimen3\font minus
  \fontdimen4\font\relax}
\providecommand{\BIBforeignlanguage}[2]{{%
\expandafter\ifx\csname l@#1\endcsname\relax
\typeout{** WARNING: IEEEtran.bst: No hyphenation pattern has been}%
\typeout{** loaded for the language `#1'. Using the pattern for}%
\typeout{** the default language instead.}%
\else
\language=\csname l@#1\endcsname
\fi
#2}}
\providecommand{\BIBdecl}{\relax}
\BIBdecl

\bibitem{Peng22}
C.~Peng, K.~Zhang, Y.~Ma, and J.~Ma, ``Cross fusion net: A fast semantic
  segmentation network for small-scale semantic information capturing in aerial
  scenes,'' \emph{IEEE Trans. Geosci. Remote Sens.}, vol.~60, pp. 1--13, 2022.

\bibitem{Zhang22}
B.~Zhang, T.~Chen, and B.~Wang, ``Curriculum-style local-to-global adaptation
  for cross-domain remote sensing image segmentation,'' \emph{IEEE Trans.
  Geosci. Remote Sens.}, vol.~60, pp. 1--12, 2022.

\bibitem{Zheng22}
A.~Zheng, M.~Wang, C.~Li, J.~Tang, and B.~Luo, ``Entropy guided adversarial
  domain adaptation for aerial image semantic segmentation,'' \emph{IEEE Trans.
  Geosci. Remote Sens.}, vol.~60, pp. 1--14, 2022.

\bibitem{Tuia16}
D.~Tuia, C.~Persello, and L.~Bruzzone, ``Domain adaptation for the
  classification of remote sensing data: An overview of recent advances,''
  \emph{IEEE Geosci. Remote Sens. Mag.}, vol.~4, no.~2, pp. 41--57, 2016.

\bibitem{Csurka2017}
G.~G.~Csurka, \emph{A Comprehensive Survey on Domain Adaptation for Visual
  Applications}.\hskip 1em plus 0.5em minus 0.4em\relax Springer International,
  2017, pp. 1--35.

\bibitem{Liu20}
W.~Liu and R.~Qin, ``A multikernel domain adaptation method for unsupervised
  transfer learning on cross-source and cross-region remote sensing data
  classification,'' \emph{ISPRS J. Photogrammetry Remote Sens.}, vol.~58,
  no.~6, pp. 4279--4289, 2020.

\bibitem{Wang21}
W.~Wang, L.~Ma, M.~Chen, and Q.~Du, ``Joint correlation alignment-based graph
  neural network for domain adaptation of multitemporal hyperspectral remote
  sensing images,'' \emph{IEEE J. Sel. Topics Appl. Earth Observ. Remote
  Sens.}, vol.~14, pp. 3170--3184, 2021.

\bibitem{Miao21}
J.~Miao, B.~Zhang, and B.~Wang, ``Coarse-to-fine joint distribution alignment
  for cross-domain hyperspectral image classification,'' \emph{IEEE J. Sel.
  Topics Appl. Earth Observ. Remote Sens.}, vol.~14, pp. 12\,415--12\,428,
  2021.

\bibitem{Liu22}
M.~Liu, P.~Zhang, Q.~Shi, and M.~Liu, ``An adversarial domain adaptation
  framework with {KL}-constraint for remote sensing land cover
  classification,'' \emph{IEEE Geosci. Remote Sens. Lett.}, vol.~19, pp. 1--5,
  2021.

\bibitem{Mateo21}
G.~{Mateo-Garcia}, V.~Laparra, D.~{Lopez-Puigdollers}, and L.~{Gomez-Chova},
  ``Cross-sensor adversarial domain adaptation of landsat-8 and proba-v images
  for cloud detection,'' \emph{IEEE J. Sel. Topics Appl. Earth Observ. Remote
  Sens.}, vol.~14, pp. 747--761, 2021.

\bibitem{Benjdira19}
B.~B.~Benjdira, Y.~Bazi, A.~Koubaa, and K.~Ouni, ``Unsupervised domain
  adaptation using generative adversarial networks for semantic segmentation of
  aerial images,'' \emph{Remote Sens.}, vol.~11, no.~11, p. 1369, 2019.

\bibitem{Li21}
Y.~Li, T.~Shi, Y.~Zhang, W.~Chen, Z.~Wang, and H.~Li, ``Learning deep semantic
  segmentation network under multiple weakly-supervised constraints for
  cross-domain remote sensing image semantic segmentation,'' \emph{ISPRS J.
  Photogrammetry Remote Sens.}, vol. 175, pp. 20--33, 2021.

\bibitem{Tasar20}
O.~Tasar, S.~Happy, Y.~Tarabalka, and P.~Alliez, ``Colormapgan: Unsupervised
  domain adaptation for semantic segmentation using color mapping generative
  adversarial networks,'' \emph{IEEE Trans. Geosci. Remote Sens.}, vol.~58,
  no.~10, pp. 7178--7193, 2020.

\bibitem{Zhu17}
J.~Zhu, T.~Park, P.~Isola, and A.~A. Efros, ``Unpaired image-to-image
  translation using cycle-consistent adversarial networks,'' in \emph{Proc.
  IEEE ICCV}, 2017, pp. 2223--2232.

\bibitem{Gatys15}
L.~A. Gatys, A.~S. Ecker, and M.~Bethge, ``Image style transfer using
  convolutional neural networks,'' in \emph{Proc. IEEE CVPR}, 2016, pp.
  2414--2423.

\bibitem{Johnson16}
J.~Johnson, A.~Alahi, and L.~Fei-Fei, ``Perceptual losses for real-time style
  transfer and super-resolution,'' in \emph{Proc. IEEE ECCV}, 2016, pp.
  694--711.

\bibitem{ref17}
X.~Mao, Q.~Li, H.~Xie, R.~Lau, Z.~Wang, and S.~Smolley, ``Least squares
  generative adversarial networks,'' in \emph{Proc. IEEE ICCV}, 2017, pp.
  2794--2802.

\bibitem{ref18}
M.~Gerke, ``Use of the stair vision library within the {ISPRS 2D} semantic
  labeling benchmark ({V}aihingen),'' 2014.

\bibitem{ref20}
T.~Miyato, T.~Kataoka, M.~Koyama, and Y.~Yoshida, ``Spectral normalization for
  generative adversarial networks,'' \emph{arXiv preprint arXiv:1802.05957},
  2018.

\bibitem{ref1}
O.~Ronneberger, P.~Fischer, and T.~Brox, ``U-net: Convolutional networks for
  biomedical image segmentation,'' in \emph{Int. Conf. on MICCAI}.\hskip 1em
  plus 0.5em minus 0.4em\relax Springer, 2015, pp. 234--241.

\end{thebibliography}

\end{document}